%% file: main.tex
\documentclass[12pt]{article} % For LaTeX2e
\usepackage{iclr2019_conference,times}
\input{math_commands.tex}

\usepackage{booktabs}
\usepackage{hyperref}
\usepackage{xcolor}		% make links dark blue
  \definecolor{darkblue}{rgb}{0, 0, 0.5}
  \hypersetup{colorlinks=true,citecolor=darkblue, linkcolor=darkblue, urlcolor=darkblue}
\usepackage{url}
\usepackage{graphicx}
\usepackage{xcolor}

\definecolor{knolcol}{rgb}{0.0,0.2,0.4}
\definecolor{humancol}{rgb}{0.0,0.2,0.4}
\definecolor{robotcol}{rgb}{0.0,0.0,0.0}
\definecolor{topiccol}{rgb}{0.0,0.0,0.0}

\def\Snospace~{\S{}} %makes section autorefs more beautiful

\title{Fillers in Spoken Language Understanding: Computational and Psycholinguistic Perspectives}

\author{Tanvi Dinkar \\
Interaction Lab\\
School of Mathematical and Computer Sciences\\
Heriot Watt University \\
Edinburgh, Scotland\\
\texttt{T.Dinkar [at] hw.ac.uk} \\
\And
Chloé Clavel \\
Social Computing Team \\
Télécom Paris\\
Palaiseau, France \\
\texttt{chloe.clavel [at] telecom-paris.fr} \\
\AND
Ioana Vasilescu \\
Laboratoire Interdisciplinaire des Sciences du Numérique (LISN) \\
University Paris-Saclay \\
Orsay, France\\
\texttt{ioana [at] limsi.fr}
}

\iclrfinalcopy
\begin{document}

\maketitle

\begin{abstract}
\emph{Disfluencies}, or interruptions in the regular flow of speech are ubiquitous to spoken discourse. \emph{Fillers} (``uh'', ``um'', \dots) are disfluencies that occur the most frequently compared to other kinds of disfluencies. Yet to the best of our knowledge, there isn't a resource that brings together the research perspectives influencing Spoken Language Understanding (SLU) on these speech events. The aim of this article is to synthesise a breadth of perspectives in a holistic way; i.e. from underlying (psycho)linguistic theory on fillers, to their annotation and consideration in Automatic Speech Recognition (ASR) and SLU systems, to lastly, their study from a generation and Text-to-Speech (TTS) standpoint. The article aims to present the perspectives in an approachable way to the SLU and Conversational AI community, and discuss what we believe are the trends and challenges in each area\footnote{This article has been published in the journal ’Traitement Automatique des Langues’ 63(3) :37-62, 2022,@ATALA. The original
manuscript is available on the web site www.atala.org}.
\end{abstract}

% \tableofcontents
% \newpage

\input{Introduction}
\input{Background/Psycholinguistic.tex}
\input{Background/Computational}
\input{Conclusion}

\section*{Acknowledgements}
Tanvi Dinkar is supported by `AISEC: AI Secure and Explainable by Construction' (EP/T026952/1), and Tanvi Dinkar and Chloé Clavel were supported by `ANIMATAS: Advancing intuitive human-machine interaction with human-like social capabilities for education in schools', European Union’s Horizon 2020 research and innovation programme under grant agreement No. 765955.

\bibliography{iclr2019_conference}
\bibliographystyle{iclr2019_conference}

\end{document}

%% file: math_commands.tex
\usepackage{amsmath,amsfonts,bm}

\def\eqref#1{equation~\ref{#1}}
\def\1{\bm{1}}

\DeclareMathAlphabet{\mathsfit}{\encodingdefault}{\sfdefault}{m}{sl}
\SetMathAlphabet{\mathsfit}{bold}{\encodingdefault}{\sfdefault}{bx}{n}

%% file: Introduction.tex
\section{Introduction}
\label{intro}
 Speech production is a complex process; i.e. ``respiratory,  phonatory,  and  articulatory  gestures are timed in such a way as to produce an acoustic signal that adequately conveys the  intended  message  both  quickly  and  smoothly'' \citep{lickley201520}. \textit{Disfluencies} can be thought of as elements that break this fluent flow of speech during speech production. Formally, they are interruptions in the regular flow of speech, such as pausing silently, repeating words, or interrupting oneself to correct something said previously \citep{Disfluency}. They occur between intentional signals (a gesture used to point at an object, \dots) and \emph{un}intentional signals (slips of the tongue, \dots) of communication \citep{corley2008hesitation}. With the increasing popularity of voice assistant technologies, an open challenge that remains is the ability to design systems that can comprehend the different signals of speech communication. However, when taking stock of a machine's capability to understand language, there is an emphasis on the learning of \emph{forms} (such as in Language Modelling (LM), where the task is string prediction), but not on \emph{meaning}; or the relationship between linguistic form and communicative intent \citep{bender2020climbing}. This issue is now widely acknowledged, for instance, the \href{https://unimplicit.github.io/}{UnImplicit workshop} state as motivation ``\dots an important question that remains open is whether such methods are actually capable of modeling how linguistic meaning is shaped and influenced by \textit{context}, or if they simply learn superficial patterns that reflect only \textit{explicitly} stated aspects of meaning \dots''. From a research standpoint, this is especially challenging when considering disfluencies, which can have \emph{implicit} and \emph{contextual} meanings -- such as when a speaker says ``uh'', ``er'' and so on.  Hence, this work is motivated by the present observations:

\paragraph{People rarely speak in the same manner with which they write.}
As \citet{bailey2003disfluencies} state, ``The processes involved in speaking and in writing differ substantially from each other, and so the products of the two systems are not the same''. The following is an example of a transcription taken from a corpus of conversational speech:

\begin{quote}
\textbf{A:} For a while there I, I, I, uh, subscribed to \textit{New York Times}, a-, actually a couple of newspapers because, uh, you know, my fiance, well, she was unemployed for a while \dots
\newline \textbf{B:} Uh-huh.
\newline \textbf{A:} \dots so she, you know, really needed to look at the, the, want-, help wanted ads.
\end{quote}

The above transcript is not \emph{fluent}, and not easily \textit{readable}. Already, some modifications have been done to the raw transcript keeping in mind the goal of readability. For instance, punctuation markers have been added to introduce sentence structure and capture the prosodic cues used by the speaker. In the transcripts, we can see that people tend to repeat themselves (``I, I, I''), interrupt each other (``Uh-huh''), rapidly shift the focus of the topic in a conversation (from newspaper subscriptions to unemployment), and are in general, \emph{disfluent}. One of the departures from written text is the presence of disfluencies. Disfluencies are frequent in speech, as fluent speech is rarely the norm; with estimates that natural human-human conversations comprise of $\approx 5-10\%$ of disfluencies \citep{shriberg1994preliminaries}. \textit{Thus disfluencies are ubiquitous to spontaneous speech.} 

\paragraph{Despite this, there are varying attitudes in the treatment of disfluencies depending on the field,} that influence research in Spoken Language Understanding (SLU). Generally, the field of psycholinguistics (dealing with communicative and cognitive aspects of language) focuses on the role disfluencies play in the production and comprehension process of speech, with many works to show their importance in speech communication. For instance, they inform us about the linguistic structure of an utterance: such as in the (difficulties of) selection of appropriate vocabulary while circumventing interruption. Other perspectives, such as a \emph{computational} one, focus on recognising disfluencies in order to remove them -- for the improvement of automatic speech recognition (ASR) systems.

\paragraph{Furthermore, ``disfluencies'' is an umbrella term used to describe a wide variety of communicative phenomena.}\label{umbrella} It is difficult to find an overarching definition of disfluency due to several confusions in terminology. \citet{lickley201520} categorises all taxonomies of disfluencies as based on \textit{form} or \textit{function}. Terminology to describe disfluencies may be based on form when the objective is to describe the ``patterns  of  words  and  syntactic  units  that  disfluencies  display''. On the other hand, function descriptions may be used when disfluencies are described with respect to the planning processes involved in speech production, deliberating over the reasons for a departure from fluency (though overlap exists between the two categories). For instance, ``uh'' may be described as an interjection (word or phrase independent of grammatical connections to other words or phrases), or a ``filled pause''/ ``filler'' (sound filling a pause in flow of speech) when considering form, or a ``hesitation'' when considering function. When ``uh'' is used as a (meta-) ``discourse marker'' (a unit of talk that brackets speech \citep{schiffrin1987discourse}), it considers both form and function -- i.e. functions to mark discourse structure, and occurs at discourse boundaries (form). Other examples of functional disfluencies can include ``self''/ ``other-repair'' (``repair'' specifically used to indicate that something has gone wrong in the planning process, and needs to be corrected). For more details of annotation used for disfluency detection tasks (which is typically form-based), please refer to \autoref{annot} Annotation for Disfluency Detection. 

\paragraph{This work is concerned with fillers} (``uh'', ``um'' \dots), compared to other disfluencies. These speech events may not contribute to the final message when considering purely a lexical level \citep{vasilescu2010role}, and thus do not have \textit{explicit} meaning \citep{meteer1995dysfluency}. Yet, they occur with \textit{high frequency} in speech datasets, compared to other structures of disfluencies. 
\citet{shriberg2001_errrr} shows that the number of fillers per word across corpora exceed other kinds of disfluencies. Additionally, they occur in an \textit{intersection between speech and text}. Fillers are a common property of spontaneous speech and are shown to have \emph{distinct acoustic characteristics/ paralinguistic properties} \citep{Shriberg1999_phonetic,vasilescu2010role}, yet they can be transcribed in text, without the requirement of more detailed annotation schemes. Thus, with the growing demand of voice assistant technologies, there is an increase in research that deals with spontaneous speech corpora -- where naturally, \textit{these tokens will frequently occur}.  

\paragraph{Yet so far, there isn't a body of work that brings together the research relevant to SLU on these specific disfluencies.} To the best of our knowledge, there is no \textit{introductory work} that brings together several perspectives surrounding fillers, i.e. from psycholinguistic to computational. While a linguistic versus computational perspective is discussed in other works \citep{lickley1994detecting}, there is a narrow focus within this perspective on disfluency detection for the purposes of removal. This ignores the broader applications of SLU that deal with \textit{social communication}. Additionally, while other linguistic works give rigorous details about the annotation of disfluencies \citep{nicholson2007disfluency,grosman2018evaluation}, including the history behind these schemes \citep{lickley1994detecting}, they are not accessible for the SLU community. This work is more targeted towards researchers in SLU and Conversational AI (or indeed, linguists interested in the work on fillers and disfluencies in SLU). Given the recent benchmark release of a filler detection task and dataset \citep{zhu2022filler}, there is a growing interest in the community on these unique disfluencies. 

Thus, \textbf{the aim of this paper} is to bring together a \textit{breadth} of perspectives on these spontaneous speech phenomena in a holistic way, considering fillers at different stages in a pipeline; i.e. from ASR to generation. Furthermore, we draw parallels in the fields of psycholinguistics and SLU; discussing research in both fields that considers fillers informative, noise \dots . Throughout the work, we discuss the challenges in each field; from issues in safety and robustness, to a lack of contextual analysis and feasibility in in-the-wild scenarios. To do so, we contrast computational approaches to disfluencies distinguished from psycholinguistic approaches, loosely adopted from \citet{lickley1994detecting}. Broadly, psycholinguistic theories deal with the communicative and cognitive aspects of language. For instance, \citet{levelt1983monitoring} studies how speakers monitor and correct their speech, and in turn, how listeners are able to integrate new material correcting the previous material. Thus in Sec. \autoref{sec: psycholinguistics} we discuss psycholinguistic research that studies the role disfluencies in the \emph{planning/production} of speech, and the \emph{comprehension} of speech. As stated, we focus on the research surrounding fillers; highlighting works that study them as informative signals of communication. However, we also present research on other disfluencies when relevant -- this is to discuss general concepts and findings from the field. Then, in Sec. \autoref{computational}, we discuss the computational perspectives on disfluencies; i.e. approaches more concerned with the \emph{recognition/processing} of disfluencies. Here, we discuss how the treatment of disfluencies can vary depending on the \textit{type of task} in SLU. Sec. \autoref{conclusion} then gives the conclusion of the article.

\textbf{This paper can be read non-linearly by referring to the legend given in \autoref{tab:my-table}}.  It is difficult to create a one-to-one mapping in the fields, given that the methodologies and approaches to each field are different. However, the intention behind some works are similar, and we believe that they can  be grouped together as complementary perspectives. 

\begin{table}[]
\begin{tabular}{@{}l|ll@{}}
              & \textbf{Psycholinguistic}                                                                                                         & \textbf{Computational}                                                                                   \\ \midrule
\textit{Production}    & \begin{tabular}[c]{@{}l@{}}\autoref{cognitive} Cognitive load\\ \autoref{communicative} Communicative Function\end{tabular}                                      & \autoref{broader_SLU} Broader SLU                                                                                  \\ \cmidrule(l){2-3} 
\textit{Comprehension} & \begin{tabular}[c]{@{}l@{}}\autoref{informative} Informative cues\\ \autoref{incremental} Results on incremental \\ processing\\ \autoref{time} Time-buying measures\end{tabular} & \autoref{generation} Generation                                                                                   \\ \cmidrule(l){2-3} 
\textit{Noisy channel} & \autoref{noise} Filtered out noise                                                                                                    & \begin{tabular}[c]{@{}l@{}}\autoref{computational} (intro) Computational \\ Perspectives \\ \autoref{sds} SLU for SDS\end{tabular}\\
\cmidrule(l){2-3} 
\begin{tabular}[c]{@{}l@{}}\textit{\underline{Note on annotation}}\\ \ref{umbrella} Introduction\end{tabular} &
    \multicolumn{1}{c}{X} &
  \begin{tabular}[c]{@{}l@{}} \autoref{annot} Annotation for \\ Disfluency Detection\end{tabular}
\end{tabular}
% \end{tabular}
\caption{Legend to show the sections of each perspective that can be read as complementary to the other.}
\label{tab:my-table}
\end{table}

%% file: Background/Psycholinguistic.tex
\section{Psycholinguistic Perspectives: from Production to Comprehension}
\label{sec: psycholinguistics}

In this section, we discuss works that study the planning/production of speech by the speaker, and the comprehension of speech by the listener\footnote{We utilise both terms ``planning'' \emph{and} ``production'', because it is not clear how intentional or unintentional disfluencies are as signals uttered by the speaker \citep{nicholson2007disfluency}.}. Please note, we consistently utilise the term ``fillers'' in this work, including when citing previous research\footnote{Please note, there are several other linguistic perspectives not discussed in this paper. Since disfluencies are ubiquitous to spontaneous speech, the literature on disfluencies is vast. For example, a socio-linguistic focus; such as studying the effect that gender, regional background, etc. have on the the production of disfluencies in \citet{shriberg2001_errrr}. Some of these perspectives, including works from psycholinguistics, \textit{do consider disfluencies as noise}. We refer the reader to the works of \citet{nicholson2007disfluency} and \citet{lickley1994detecting}, which give a comprehensive overview of these perspectives.} that may use other terms to describe the same phenomena\footnote{\citet{Clark2002_using} introduced the term ``fillers''; as the term ``filled pauses'' seemed to indicate that there is a pause in speech \emph{filled} by some sort of (meaningless) sound.}.

\subsection{Production} 
\label{ssec: production}

\subsubsection{Cognitive load}\label{cognitive} A common theme in the planning process is the idea of \emph{cognitive load}, i.e. the amount of cognitive \emph{effort} required in the planning process. The production of disfluencies resulting from the planning process has been considered at different linguistic levels. For example, a prosodic/acoustic analysis found that speakers tend to maintain a fixed speaking rate during most utterances, but often adopt a faster or slower rate, depending on the cognitive load \citep{o1995timing}. From this, it was found that disfluent speech may be a result of cognitive load; i.e speakers slow down their speech when having to make unanticipated choices (for example, using more fillers), and accelerate their speech when repeating some words. This shows that the \emph{types of disfluencies produced} may give further insight to this planning process. The link between the rate of speech and the different disfluencies produced was also found in \citet{shriberg2001_errrr}. Two groups of speakers were identified -- \emph{repeaters}, who produce more repetitions (when what was said is exactly repeated) than deletions (when previous material that was uttered is abandoned), and \emph{deleters}, who produce more deletions than repetitions. The \emph{repeater-deleter} difference was not only due to stylistic variation in speakers; deleters were found to have a faster speaking rate than repeaters in terms of words per unit time. The interpretation suggested here (contrary to \citet{o1995timing}) is that speakers with a slower speaking rate (repeaters) ``take more time to plan'', leading to an increase in repetitions, while faster speakers (deleters)  ``get ahead of themselves'', and recant what was said to begin again. The different conclusions could be due to the \textit{difference in dataset size, context and domain}\footnote{Indeed, \citet{shriberg2001_errrr}'s findings were based on conversational style dialogues (human-human) in addition to task-oriented dialogues, while \citet{o1995timing} focused on task-oriented dialogues (human-machine).}. For instance, speakers have been shown to be more disfluent in dialogues compared to monologues \citep{oviatt1995predicting}, in human-human conversations than human-machine conversations \citep{oviatt1995predicting}, and disfluencies are affected by dialogue role and domain \citep{colman2011distribution}.
% ; . 

The analysis of disfluencies at other levels also reveal \textit{characteristics of the planning process}. In an acoustic-syntactic analysis, it was found that high-frequency monosyllabic function words (such as ``the'' or ``I'') are more likely to be prolonged or have a fuller form when there are neighbouring fillers (``uh'' and ``um''), indicating that the speaker was encountering problems in planning the utterance \citep{bell2003effects}. At an utterance level, \citet{shriberg2001_errrr} found that the longer the utterance, the more disfluencies they contain, also suggesting an increase in cognitive load of the speaker. Disfluencies were also found to occur at the start of an utterance, due to higher cognitive load in planning an utterance \citep{maclay1959hesitation}. At a discourse level, \citet{swerts1998filled} found that fillers can be used by the speaker to indicate pausing to (re)formulate thoughts, particularly at discourse boundaries. 

\subsubsection{Communicative function}\label{communicative} \citet{beattie1979contextual} suggested that instead of cognitive load, there is an \emph{element of speaker choice} in the planning process. They first establish that generally, speakers are disfluent both when producing low frequency words and improbable words in the context, focusing on fillers ``ah'', ``er'', and ``um''. However, even when frequency of a word was maintained by the speaker, they were still disfluent when producing words with low contextual probability. Like this, there are two main positions behind the speaker's production of disfluencies. One is that disfluencies are accidentally caused in speech due to \emph{cognitive burden} of the speaker (such as \citet{bard2001disfluency}). Other works study disfluencies as an important \emph{communicative function} used in dialogue. This view is based on the \emph{strategic modelling view}, where the speaker strategically updates the listener, by using disfluencies as cues \citep{nicholson2007disfluency}. The distinction between the two views is that the former is an unconscious by-product of speech produced by cognitive (over)load, while the latter is an intentional and strategic production by the speaker. Often studies will look at both of these positions, by analysing the individual disfluencies of a speaker as well as the collective disfluencies produced by interlocutors. The results for the production of disfluencies are often mixed, such as in \citet{nicholson2007disfluency} and \citet{yoshida_disfluency_2010}, with evidence suggesting that they occur in both cases; i.e. speakers may both strategically and unconsciously produce disfluencies depending on the context. 

\subsection{Comprehension}
\label{ssec: comprehension}
Research also focuses on the \emph{comprehension} of disfluent speech, i.e. taking into account the listener's understanding of the speaker's disfluencies \citep{corley2008hesitation}, and not on why the disfluency itself was produced \citep{nicholson2007disfluency}.  As \citet{corley2008hesitation} state, ``it is hard to determine the reason that a speaker is disfluent, especially if the investigation is carried out after the fact from a corpus of recorded speech''. Works that study the effect of disfluencies on listener comprehension state that listeners \emph{must} have developed a comprehension system to process disfluencies, given how frequent they are in spoken language. 

\subsubsection{Informative cues}\label{informative} Research shows that listeners can use disfluencies as signals to \emph{understand and resolve incoming information} in the flow of speech, regardless of whether the speaker intentionally used disfluencies in that way. Consider the following example taken from \citet{Brennan1995_feeling}:
\begin{quote}
\textbf{A:} Can I borrow that book?
\newline \textbf{B:} \dots \{F um\} \dots all right.
\end{quote}

Here, speaker \textbf{B} used a filler \{F...\}, which causes \textbf{A} to note that \textbf{B} might have had a different intention compared to if \textbf{B} answered ``all right" immediately. While this example is on a pragmatic level (i.e. the listener notes the discrepancy between what was said in essence, and how it was said), works also suggest that listener's can use disfluencies as \textit{communicative cues} in other ways.

For instance, it was found that fillers helped in the faster recognition of a target word for listeners, indicating that they cause listeners to pay more attention to the upcoming flow of speech \citep{tree1995effects}. The use of fillers also \emph{biases} listeners towards new referring expressions rather than ones already introduced into the discourse \citep{arnold2004old}. \citet{arnold2007if} additionally showed that listeners have expectations on the upcoming material to contain \emph{difficult to describe}/unconventional referring expressions when preceded by the filler ``uh''. Listeners expect the speaker to refer to something new following the filler ``um'', compared to noise of the same duration (such as a cough or sniffle) \citep{barr2010role}. This result was found to be \emph{speaker specific} for the listener. This means that the listener was able to consider what was an already introduced referring expression versus a new referring expression for the current speaker; not just what was old or new for themselves (the listener). \citet{barr2010role} suggest that this is evidence for the \emph{perspective taking account of language comprehension}. This account suggests that listeners are able to interpret fillers as delay signals, and then infer on plausible reasons for this delay in speech by considering the speaker's perspective. This implies that fillers can have a \emph{metacognitive effect} (i.e. assessment of knowledge state), with the listener using fillers as cues to interpret the speaker's metacognitve state. Fillers and prosodic cues were also found to impact listener's attributions of a speaker’s metacognitive state, specifically the estimation of a speaker's level of certainty on a topic \citep{Brennan1995_feeling}. 

\subsubsection{Filtered out noise}\label{noise}
As a counterpoint to the works discussed, research also suggests that disfluencies may be perceived as noise and thus filtered out by the human listener. In experiments, \citet{tree1995effects} noted that removing repetitions from the utterance digitally did not affect the rating of perceived naturalness of speech. However, this could be explained from later work by \citet{Shriberg1999_phonetic}, who found that repetitions have similar pitch contours, but just stretched out over time. \citet{lickley1991processing} found that listeners could not pinpoint the exact interruption point of a disfluency, and tended to point it out later (up to one word) in the flow of speech.  \citet{lickley1998can} studied a listener's ability to identify (several types of) disfluencies to find that they are not reliably predictable unless a noticeable pause or abandoned word is apparent. Note, the works discussed here on disfluencies as cues to integrate information are mainly targeted towards fillers, compared to these works that study more complex disfluencies\footnote{Confusions in terminology arise from research itself, as "disfluencies" is an all-encompassing term used for many works, including when the works are \textit{only} concerned with fillers.}.

\subsubsection{Results on incremental processing}\label{incremental}
\citet{bailey2003disfluencies} point out that the idea of ``filtering'' out disfluencies (despite their prosody remaining intact, location not easily remembered by listeners) does not account for the \emph{incremental} nature of speech processing; i.e. the processing starts before the input is complete. Filtering would imply that processing needs to occur \emph{after} the removal of disfluencies, and that the listener then, would need to wait for the entire utterance to be completed by the speaker before processing. However, the processes involved in comprehension are continuous and \emph{incremental}, as humans process utterances incrementally. Listeners \emph{must} have developed a comprehension system in order to process disfluencies, given how frequently they occur in spoken language \citep{bailey2003disfluencies}. Disfluencies are part of the incremental processing of the flow of speech; for instance \citet{bailey2003disfluencies} show that disfluencies can affect the internal syntactic parser of the listener. The disfluencies considered here was the filler ``uh'', but also any kind of interruption (noise) was also deemed to be a disfluency. \citet{brennan2001listeners} show that listeners may use disfluencies as cues to avoid integrating what they deem to be incorrect material in an online processing task. Thus, there is evidence to support that fillers are not filtered out by the listener, and indeed, included in the \textit{online processing of the utterance}.

Fillers used as cues to understand new information has even been shown \emph{neurologically}, and in an online processing task. \citet{corley2007} studied the effect of filler (``um'', called a ``hesitation'' in the work) on the listener’s comprehension using the $N400$ function of an Event-related potential (ERP). The $N400$ effect can be observed during language comprehension, typically occurring $400$ ms after the word onset; it is a negative charge recorded at the scalp consequent to hearing an unpredictable word. The $N400$ effect was first established in listeners who heard unpredictable words compared to predictable words. Then, when the filler proceeded the unpredictable word, the $N400$ effect in listeners was \textit{visibly reduced}. In a subsequent memory test on the listener, words preceded by this filler were more likely to be remembered. 
% new research from venderbelt 

\subsubsection{Time-buying measures and Effects on Memory}\label{time} Works also hypothesised whether all these effects of integration can be explained by the \emph{processing time hypothesis}. This means, considering whether the disfluent speech is more memorable/noticeable simply because disfluencies add more time to the speech utterance (referring to research that studies the role of fillers in giving pause to the discourse). Thus, does this effect -- i.e. a pause in the utterance -- cause the listener to simply give more attention to the utterance? \citet{Fraundorf2011_Attentional} examined this in a study on how fillers affect the memory of the listeners. They exposed the listeners to fluent speech containing fillers versus coughs of equal duration artificially spliced into the speech. They found that while fillers facilitated recall, therefore being beneficial on memory, coughs negatively hampered recall accuracy. Disfluent speech (fillers) is hence more likely to be remembered by the listener, and this is \textit{not solely based on the additional time added to the utterance}. \citet{Fraundorf2011_Attentional} also manipulated the location of the fillers in speech to study the effect of \emph{position} of fillers on comprehension. This was based on the findings of \citet{swerts1998filled} (discussed previously), who found that following fillers, listeners may expect a speaker to shift topics as they carry information about larger topical units -- therefore, acting as cues for discourse structure. However, \citet{Fraundorf2011_Attentional} found that fillers benefit listener's recall accuracy regardless of its typical or atypical location. \citet{Tottie2014_Use} found that fillers are noticed more when overused or used in the wrong context, so while they may facilitate recall regardless of location, they still may be more noticeable in atypical locations.

Thus the speaker produces disfluencies, and we have illustrated works that show that they may be strategically used, but also can indicate problems in the planning process. When comparing production versus comprehension, regardless of whether the speaker intentionally uses disfluencies \citep{barr2010role}, the listener still \emph{integrates} them as cues for upcoming information. However, it seems that the type of disfluency may also matter in the process of comprehension and whether the listener integrates these disfluencies or filters them out. 

%% file: Background/Computational.tex
\section{Computational Perspectives: Removal versus Integration}
\label{computational}
We briefly defined the computational perspective on disfluencies; i.e. works concerned with the computational processing of disfluencies in order to remove them; rendering the utterance more ``fluent'' and text like. 
In this distinction, we also consider linguistic work motivated by this goal. For example, phonetic works that characterise the acoustic environments that disfluencies occur in, with the ultimate goal of aiding Automatic Speech Recognition (ASR) systems. While it is a general trend to consider \emph{all} disfluencies as noise in this perspective, we also distinguish research that has found disfluencies informative in SLU. SLU is broad and not one defined task, it ``combines speech processing and natural language processing (NLP) by leveraging technologies from machine learning (ML) and artificial intelligence (AI)'' \citep{tur2011spoken}.
Thus we distinguish between what is generally considered SLU (SLU for Spoken Dialogue Systems), and broader SLU (SLU tasks concerned with social communication), loosely adopting the distinction made in \citet{tur2011spoken}. We make this distinction, because the treatment of disfluencies varies greatly -- from \emph{removal to integration} -- depending on the SLU task. 

\begin{figure}[h]
\begin{center}
\includegraphics[width=\textwidth]{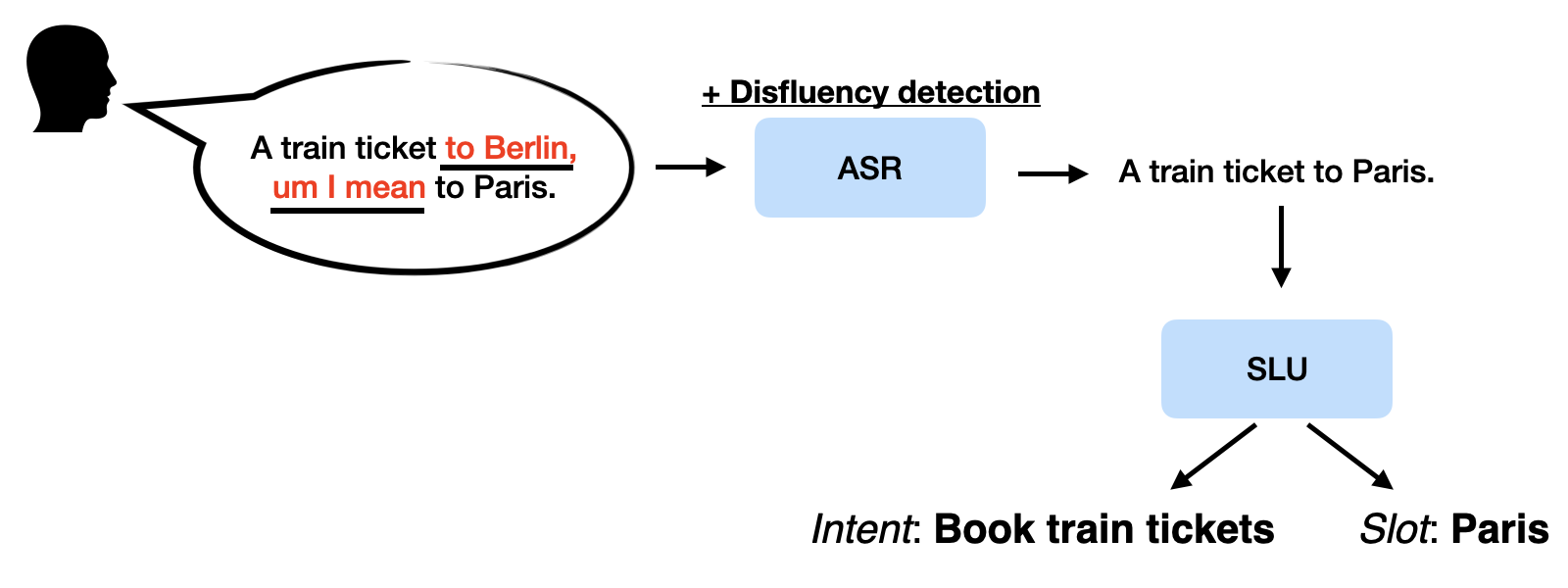}
\caption[SLU for SDS example]{Example of SLU for SDS, where the disfluent part of the utterance is highlighted in red and underlined. As shown, the disfluencies are removed in post-processing using disfluency detection systems, after ASR systems transcribe the input speech. Then, the input is further collapsed into a semantic frame, consisting of intent and slot.}
\label{fig: intent}
\end{center}\vspace{-0.7cm}
\end{figure}

\subsection{SLU for Spoken Dialogue Systems (SDS)}
\label{sds}
In SLU for SDS, the objective is to collapse the input utterance into a \textit{semantic frame}, consisting of an \emph{intent} and a \emph{slot} \citep{louvan2020recent}. Consider the disfluent utterance ``A train ticket \underline{to Berlin uh I mean} to Paris''. If ``\underline{to Berlin uh I mean}'' is not removed (marked in red in \autoref{fig: intent} to get the sanitised utterance; ``A train ticket to Paris''), it leads to confusion in the subsequent (semantic) processing of the utterance -- the correct intent would be identified (i.e. book train tickets) and but not the correct slot (i.e ``Paris'' not ``Berlin''). Disfluencies in this regard are discarded as noise. Thus in a standard SDS pipeline, the output transcripts of ASR systems are cleaned of disfluencies in post-processing as shown in \autoref{fig: intent}, using disfluency detection systems. From a computational perspective, the intent to process of disfluencies in present-day is due to automatic Natural Language Understanding (NLU) systems not being \textit{robust} to them \citep{Ginzburg14disfluenciesas}. Representations of this semantic frame focus on the lexical aspect of the utterance but neglect the non-lexical channel. Thus removing disfluencies such as fillers could remove important information about the \emph{social} aspects of communication (subsequently discussed in \autoref{broad} Broader SLU and Generation). 

There has been an interest in studying the characteristics of disfluencies for recognition purposes, such as how they are distributed in corpora (for instance, \citet{shriberg1994preliminaries} and \citet{shriberg2001_errrr}), and also, at different linguistic levels. The motivation of these works was to identify the properties of disfluent speech to \emph{enable a comparison with its fluent counterparts}. For instance, at a phonetic level, \citet{Shriberg1999_phonetic} found that the vowels in fillers gave much longer durations than the same vowels in fluent contexts, and \citet{shriberg1993intonation} found that the intonation of fillers is not independent of prior prosodic context. These studies were done with the aim of ideally \emph{leveraging these features for ASR to be more robust to  disfluencies}, but with mixed results (such as in \citet{shriberg1997prosody}). Other linguistic levels have also been studied, such as morpho-syntactic features \citep{goryainova-etal-2014-morpho}, contextual occurrence \citep{vasilescu2010role}, pragmatic levels \citep{shriberg1998can} and so on.  

\subsubsection{Disfluency Detection}\label{dis_detection}
\citet{johnson2004tag} were the first to introduce the task of disfluency detection, framing it as a noisy channel problem; i.e. noise (disfluencies) has been added to the fluent source utterance. Presently, the task of detection is framed in many ways. For instance, as a \textit{sequence-tagging} task, where typically each token in the input utterance needs to be classified as either in the beginning, inside, or outside (BIO) of a disfluent region. This has been adapted to neural architectures, for instance \citet{zayats2019giving}, with a bidirectional Long Short Term Memory (biLSTM) network, or a Transformer model from \citet{jamshid-lou-johnson-2020-improving}. Encoder-decoder frameworks allow for detection to be framed as a \textit{sequence-to-sequence neural translation task}, where the encoder learns a representation of the disfluent utterance, and the decoder generates a fluent version of that utterance (see \citet{wang-etal-2018-semi}). Similar to this, the objective of end-to-end \textit{speech translation} is to output fluent text from an input of disfluent \textit{audio}. This frames the detection problem as an interim step between ASR and a downstream task such as machine translation (see \citet{salesky-etal-2019-fluent}). \citet{jamshid-lou-johnson-2020-end} propose an end-to-end speech recognition and disfluency removal system, so that the fluent output may be used for any downstream task. \citet{hough2017joint} propose a \textit{joint task} of disfluency detection and utterance segmentation. They find that the task is better approached jointly, with the combined system outperforming results on individual tasks. Detection may be combined with other tasks in a multi-task learning (MTL) scenario, showing improvements when there is a secondary task of language modelling \citep{igor2018dis}, or even simultaneous tasks of detection, POS-tagging, language modelling and utterance segmentation \citep{houch_2020}. 

A drawback on state-of-the-art (SOTA) systems is they are not robust to transcribing disfluencies correctly in-the-wild, with criticisms that they could not work in an online processing task \citep{google_disfluency} nor with raw transcripts from ASR \citep{rohanian2021best}. Specifically for \textit{disfluency detection}, often the \textit{textual input} used is conditioned on already appropriate segmentation and accurate transcription -- which is unrealistic in a real-world scenario. From \citet{rohanian2021best}, works on disfluency detection ``are almost \emph{exclusively} conducted on pre-segmented utterances of the Switchboard (SWBD) \citep{godfrey1992switchboard} corpus of telephone conversations''; contradicting the generalisation capabilities of such systems. Works are exploring the feasibility of detection with online processing \citep{google_disfluency,rohanian2021best,igor2018dis,hough2017joint}, with transcripts arising from ASR \citep{rohanian2021best}. With hybrid conversations, where user interactions switch between task-oriented and open domain requests \citep{kim2021robust}, this topic of detecting disfluencies is starting to come to the forefront once more. The improvement of these systems would be beneficial for other types of research, i.e. better automatic annotation of disfluencies in corpora in order to study disfluent phenomena -- particularly for areas such as clinical SLU, to help determine a patient's cognitive state.

\subsubsection{Filler Detection}\label{filler_detection}
% \textbf{TODO}: mention the google paper, the balance between latency and accuracy, only a handful of studies working on online processing. 
Systems may be better adapted to recognising specific kinds of disfluencies depending on their properties; fillers for instance, are acoustically distinct. Off-the-shelf open-source speech recognisers such as CMU Sphinx \citep{lamere2003cmu} provide functionality to define a filler phone dictionary of \textit{paralinguistic sounds} that are not present in the Language Model (LM). \citet{das2019increase} leverage the acoustic properties of fillers by training a convolutional recurrent neural network (CRNN) applied directly to \textit{audio recordings} to do filler segmentation, i.e. segmenting fillers in order to output a more fluent utterance. Fillers here are considered to be ``unprofessional speech'', stating that the speaker sounds more ``fluent, confident and practised compared to the original recorded speech'' evaluated by automatic measures\footnote{Proposed in \citet{kormos2004exploring} to study the perception of fluency in second language learners. Indeed they state in the work that the number of fillers \textit{did not affect the perception of fluency the raters had} (see drawbacks discussed in \autoref{broad} Broader SLU).}, rather than human annotation judgements. Regardless, it would be beneficial to many branches of research to automatically label fillers reliably. However, as pointed out in \citet{das2019increase}, these systems can suffer from false positives (for instance the ``um'' in ``umbrella''). Recently, a new filler detection benchmark and dataset was introduced, where the objective was to label filler candidates given an incoming speech input \citep{zhu2022filler}. The system leverages voice activity detection (VAD) and ASR to detect possible filler candidates from other tokens, and then a classifier to categorise them. They evaluate the ASR approach to find it outperforms a keyword spotting approach. The publicly available dataset is based on podcast episodes and contains a diverse range of speaking styles and topics -- which could facilitate many future research directions. Thus, there is attention now given to individually detecting fillers from raw audio as a task, without gold-standard transcripts.

Interestingly, while Google Cloud Speech-to-Text services map paralinguistic sounds such as fillers to silence, it has introduced a functionality in its \href{https://cloud.google.com/speech-to-text/docs/medical-models?hl=en}{medical ASR model} to include fillers when transcribing medical documents. To contrast, in the \href{https://cloud.google.com/dialogflow/cx/docs/concept/agent-design}{Google Dialogflow API}, there are guidelines given in order to \textit{avoid} fillers as giving unnecessary variety to utterances, as they are ignored by subsequent NLU models (that collapse the input into a semantic frame as shown in \autoref{fig: intent}). What is interesting here is the \textit{exclusion} of fillers in an SLU for SDS task (i.e. Google Dialogflow), and the \textit{inclusion} of fillers in an ASR task with a broader SLU context -- i.e. to pick up social cues in medical transcription. Thus the treatment of fillers and disfluencies is evolving in SLU research to consider specifically the nature of the task; compared to the previous view that all disfluencies must be removed as noise. Hence detection may focus on \textit{removal} if the subsequent task is to collapse the input into a semantic frame, or \textit{integration} if the social aspects of communication may be further considered. 

\subsubsection{Annotation for Disfluency Detection}
\label{annot}

Annotation schemes for disfluency detection are almost always form based, as the intent is to sanitise the utterance of disfluencies. For this, there needs to be precise, form-based characterisations of the environment in which disfluencies occur, with no interest in deliberating as to why there was a departure from fluency. We briefly outline the most commonly used annotation scheme in disfluency detection, consistent with the Switchboard repair mark-up \citep{meteer1995dysfluency}. Consider the following example:

\begin{equation}
\text{Archie\;\;}{\underbrace{\text{[\;likes\;}}_{\text {RM}}} + {\underbrace{\text{\{F\;uh \} }}_{\text{IM}}}  {\underbrace{\text{\;\;loves\; ]}}_{\text{RP}}} \text{\;\;Veronica} . 
\label{eq: standard_eg}
\end{equation}

At a high level, there is the notion of \emph{some kind of erroneous speech that is to be replaced by corrected speech} -- here,``loves'' to replace ``likes''. This is used as a starting point for annotation (originally proposed by \citet{levelt1983monitoring}). While variations of this scheme exist, the underlying reasoning \textit{specifically} for detection tasks remains the same. Formally, there is: \romannumeral 1) the \textit{reparandum phase} ($RM$), i.e\ or the entire region to be deleted, and \romannumeral 2) the \textit{repair phase} ($RP$), i.e.\ what replaces the $RM$. This was adopted by \citet{shriberg1994preliminaries}, who also proposes the term \romannumeral 3) \textit{interregnum phase} $IM$, which is an optional interruption point, where the speaker may realise that a correction needs to be made. \textit{Non-sentence elements} (such as fillers) can occur within this $IM$ structure or outside, also called \textit{isolated edit terms/ \textit{single edit tokens}}. From this structure, several types of disfluencies can be formally defined. For instance \textit{repetitions} -- i.e. when the $RM$ phase is repeated exactly in the $RP$ phase, and \textit{restarts} -- i.e. when the $RM$ phase is discarded. This annotation structure is useful for disfluency detection and SLU because the reasoning is to only keep the corrected speech in the utterance (i.e. the $RP$) for downstream processing (see \autoref{fig: intent}), and discard the rest as noise ($RM$+$IM$).

However, a drawback of this scheme for a detection task is that it spans only one speaker turn and can only be initiated by the speaker producing the disfluencies. In many contexts, repair could span different speaker turns and different interlocutors (see a detailed discussion in \citet{purver_computational_2018}). Disfluency detection thus can fail on longer disfluencies \citep{Zayats2019_disfluency}, and disfluencies spanning multiple turns \citep{purver_computational_2018}. Additionally, many other annotation schemes exist when the objective is not constrained to detection alone. For instance, \citet{christodoulides-etal-2014-dismo} proposed a tool to holistically do a form-based annotation of several phenomena characteristic of spontaneous speech: with disfluency detection and annotation, and multi-word unit recognition (including POS, syllable, phone tagging \dots). Rather than an annotation specific to one task, these layers of annotation could be beneficial for several downstream SLU tasks. The annotation scheme is more realistically designed for spoken language -- such as using prosodic cues for segmentation in the absence of punctuation. There are also language specific schemes proposed (such as \citet{benzitoun-etal-2012-tcof}, \citet{kahane2020annotation} and \citet{eshkol-etal-2010-eslo} for French), as \textbf{a limitation of this survey is that we constrain ourselves to English}\footnote{Note also that the issue of annotation has been extensively studied from a non-computational linguistic perspective \citep{grosman2018evaluation}, with extensive summaries of terminology from different linguistic works \citep{lickley1991processing,nicholson2007disfluency,lickley201520}, and new and emergent annotation schemes proposed \citep{crible2015annotation} which are not discussed here.}. 

Problems may also arise when transcribing (disfluent) audio itself. \citet{Le2017_stance} found that transcribers who had transcribed a few conversations tended to have \textit{substantially higher transcription errors}, compared to transcribers that had transcribed a large number of conversations. \citet{Zayats2019_disfluency} found that tokens generally related to spontaneous speech phenomena (fillers \dots) have a high frequency of transcription errors, which they \textit{discuss could be due to these tokens being non-standard}, i.e. unaccounted for in annotation instructions. Fillers in particular, are among some of the most likely tokens to be mis-transcribed (whether inserted, deleted or substituted), and transcriber experience has a noticeable effect on the accuracy of transcribing fillers specifically \citep{Le2017_stance}.

\subsubsection{Text Representations of Disfluencies}\label{representation}
With the increasing popularity of voice assistant technologies, a trend has emerged in Natural Language Understanding (NLU) research to overlap with SLU; i.e. considering the \emph{textual processing of speech transcripts} (as discussed in \citet{Ruder_2020} as ``Speech first-NLP''). However, discrepancies may arise in the processing of speech transcripts compared to processing grammatically written text, if utilising the same (NLP) systems.  For example, \citet{tran2017joint} present an attention-based encoder-decoder model for parsing conversational sentences arising from speech transcripts. An important finding from this work was that the integration of acoustic-prosodic features showed \textit{the most gains over disfluent and longer sentences} compared to fluent ones. This empirically \textit{shows a discrepancy} between transcripts that are more ``speech-like'' (i.e. are disfluent and have acoustic variation) compared to transcripts that are more grammatical and resemble written text. It is worth noting that for this task of parsing, there is the assumption of \emph{already structured} spontaneous speech transcripts to be used as input (i.e.\ annotated punctuation, input being pre-segmented sentences \dots) -- so indeed, there is a further gap between perfectly fluent, grammatical text versus raw ASR/verbatim transcripts. \citet{herve-etal-2022-using} investigate how deep contextualised embeddings could be pre-trained on massive amounts of ASR generated transcripts for more realistic samples in spoken language modelling. The downstream tasks including parsing conversational utterances show improvements of the model pre-trained on raw ASR data (which they call FlauBERT-Oral) compared to the original model trained on written data (i.e. FlauBERT, a French LM \citep{le-etal-2020-flaubert-unsupervised}). They point out that the ASR generated only lowercase transcripts, and that adding capitalisation and punctuation to the transcripts could benefit the model i.e. \textit{re-introducing some degree of sentence structure}.

\citet{Tran2019} showed that deep contextualised embeddings pre-trained on large written corpora can be fine-tuned on smaller spontaneous speech datasets to improve parsing on conversational speech transcripts. While this indicates that some \textit{general} characteristics of spontaneous speech may be learnt in the fine-tuning stage (and as shown by \citet{herve-etal-2022-using} using models pre-trained on raw ASR), results are mixed when specifically considering the text representation of fillers. \citet{valentin2017} for instance showed that pre-trained word embeddings such as Word2vec \citep{word2vec} have poor representation of spontaneous speech phenomena such as ``uh'', as they are trained on written text and do not carry the same meaning as when used in speech. Interestingly, this discrepancy may not always negatively impact the \textit{robustness} of the system when purely considering fillers. For instance, \citet{dinkar-etal-2020-importance} investigated the representations of fillers using deep contextualised word embeddings. They found that Bi-directional Encoder Representations (BERT) \citep{devlin2018bert}, has existing representations of fillers, despite being pre-trained on massive amounts of written text. They found (similar to \citet{stolcke1996statistical}'s results on $ngrams$) that the inclusion of fillers \textit{reduces} the uncertainty of a language model in a spoken language modelling task. This is despite research that shows that overall, speech disfluencies occur at higher perplexities \citep{sen-2020-speech}. Thus, in addition to psycholinguistic work, there is computational research to show that indeed, fillers may be able to provide information regarding the neighbouring words to the right in a language modelling task. However, \citet{dinkar-etal-2020-importance} found that BERT is unable to distinguish between the two fillers ``uh'' and ``um'', despite research to show that they occupy different functions in discourse \citep{Le2017_stance,dinkar2022computational}. To conclude, there is now the general awareness that spoken speech is not like written text, and increasing studies work on how to learn representations of spontaneous speech given that architectures are usually developed on large amounts of written data. 

\subsection{Broader SLU and Generation}
\label{broad}
 
\subsubsection{Broader Spoken Language Understanding (SLU)}\label{broader_SLU} We consider broader SLU as the analysis of the flow of speech by leveraging NLP and ML techniques, (loosely adopted from \citet{tur2011spoken}) -- particularly for higher order tasks that are concerned with social/affective computing. Though the methodologies differ from the psycholinguistic approach to study the production contexts of disfluencies, the underlying intent is similar. That is, to study the context of disfluencies produced (and indeed, many other lexical and non-lexical features) and their link to a variety of socio-communicative and cognitive phenomena. The findings are useful for feature engineering in broader SLU tasks. For example, disfluencies in several works have been found to be an informative \emph{social signal} \citep{ekman1980relative,mairesse2007using,vinciarelli2014survey,schuller2019affective}.
Fillers specifically, are commonly used as an attribute to study big 5 personality traits \citep{mairesse2007using}. The Computational Paralinguistics Challenge focused on detecting fillers, which they considered to be a ``social signal'' \citep{schuller2019affective}, acknowledging the importance of fillers in \textit{tasks concerned with social communication}. Along these lines, research in personality computing has the most consistent correlation, with observations made from speech; including \textit{paralanguage} such as fillers \citep{ekman1980relative,vinciarelli2014survey}. We cannot exhaustively account for these works, as they are numerous. Some examples are research on the role of disfluencies (mainly fillers) in the prediction of a speaker's \textit{emotions} \citep{moore2014word,tian2015emotion}, \textit{stance} \citep{Le2017_stance}, perceived \textit{persuasiveness} \citep{park2014computational}, perceived \textit{confidence} \citep{dinkar-etal-2020-importance}, etc\dots \citet{dufour2014characterizing} point out that spontaneous speech is not the same as prepared speech, where the utterances are well formed and closer to written documents. They focus on classifying the \textit{degree of spontaneity} of speech in order to do SLU tasks, such as characterising speaker roles (example, an interviewer versus an interviewee) using fillers and other acoustic-linguistic features.  Recent work argues that disfluencies will be useful in dialogue based computer-assisted language learning; i.e. detecting and analysing a learner's disfluencies (including silences and laughter) could potentially help a system determine appropriate pedagogical interventions \citep{skidmore-moore-2022-incremental}. Disfluencies may play a crucial role in \textit{clinical SLU} tasks, such as in dementia recognition \citep{rohanianalzheimer}. As stated, Google Cloud Speech-to-Text has included in its medical ASR model options to include fillers, for the purpose of transcribing medical documents.

 However, an open challenge that remains in analysis is the inclusion of context. Research may predetermine based on the scenario whether disfluencies are positive or negative. For instance, in the automatic processing of job interview data \citep{7838991}, fluency is assumed to be desirable. While indeed, the speaker's production of disfluencies may have an effect on the outcome of a job interview; there are nuances of how the speaker utilises such spontaneous speech phenomena; i.e. both production and perception will vary based on socio-linguistic background, context, domain and so on. Analysis may be based on the idea that the frequency of words (including, non-lexical tokens such as fillers present in transcripts) can represent underlying affective traits \citep{boyd2021natural}. In this aspect, the link between fillers and a wide variety of phenomena is to be expected -- from linguistic levels to higher affective levels; with \citet{barr2001trouble} even describing fillers as \emph{vocal gestures}. In a recent survey, \citet{boyd2021natural} discuss this drawback regarding research in the intersection between psychology and language analysis -- but also considering interdisciplinary fields such as social computing; stating that it may not always be a case of ``paying attention to $X$ is correlated with $Y$''. Thus we should cautiously assume a linear relationship between the fillers produced and the task under consideration. \citet{dinkar2021local} for instance found that speakers tend to stylistically use fillers before introducing new information in the discourse, but that listeners may not associate this specific use of fillers with their estimation of the speaker’s confidence. 
 
An additional challenge is the varied terminology used to describe fillers in the context of broader SLU. While standardised terminology may be used in a benchmark task such as detection, it is not the case for less standardised tasks. This is problematic, because it can lead to a lack of availability and transparency of the findings. For instance, the \emph{Linguistic Inquiry and Word Count (LIWC)} \citep{pennebaker2001linguistic}, a commonly used text analysis software in personality computing \citep{mehta2020recent}, gives guidelines to annotate ``nonfluencies'' (fillers) -- ``uh'', ``um'', ``er'', and ``stuttering''. Here, ``stuttering'' broadly denotes the general phenomena of being disfluent\footnote{Please note, the term is not to be confused with the clinical sense of ``stuttering''. In clinical literature, the term commonly used is ``stutter-like disfluencies'' (SLDs). This use seems borrowed from \citet{mahl1956disturbances}, who used the term to refer to repetition of partial words \citep{lickley201520}.}. 

\subsubsection{Generation}\label{generation}  Similar to psycholinguistic approaches, it is to be noted that there are works that study the \emph{perception} of disfluencies from a generation standpoint, i.e. using artificially synthesised disfluencies in speech. While this is not under the umbrella of SLU as such, we briefly describe some research specific to fillers. For instance, in a work directly motivated by psycholinguistic perspectives of comprehension, \citet{wollermann2013disfluencies} explore the listener’s perception of disfluencies using Text-to-Speech (TTS). This was based on the work of \citet{Brennan1995_feeling}, that discusses the role of fillers and prosodic cues in a listener’s evaluation of how uncertain they think the speaker is regarding a topic. They had the system exhibit ``uncertain'' behaviour through disfluent TTS responses in a question-answering context. They found that disfluencies in combination with prosodic cues (i.e. delays + fillers) increased a listener’s perception of uncertainty towards the system’s answers. Similarly, \citet{kirkland2022s} found that when fillers were not present in synthesised speech, it lead to a perception of more confident sounding utterances, while utterance-medial fillers lead to a perception of the least-confident sounding utterances (both in addition to other prosodic features). 

Some works may not have basis in psycholinguistic theory. For instance, there is research that considers how disfluencies enhance the \emph{naturalness} of the synthesised speech. \citet{pfeifer2009should} evaluate an agent that uses fillers ``uh'' and ``um'' in
speech.  The motivation behind this was to improve the naturalness of speech in an Embodied Conversational Agent (ECA), as ECAs often try to emulate humans in gestures and facial expressions, yet speak in fluent sentences. Results are mixed, with some participants saying that fillers enhanced the naturalness of the conversation, while  others expected that an agent should speak fluently, and fillers were deemed inappropriate. \citet{szekely2019train} discuss approaches for treating fillers in TTS tasks, i.e. suggesting methods that will result in them being synthesised naturally (both distributionally and perceptually) in the generated output. Disfluencies may also be utilised as a \emph{communicative strategy} in generation. \citet{skantze2015exploring} studied how a system can use multi-modal turn-taking signals (including fillers) as a \emph{time-buying measure}, i.e. to buy time for generating a response as the next move of the robot is decided. 

While a common goal of AI is to work towards more human-like (anthropomorphic) agents, a challenge that remains is to consider the trade-off between the naturalness of a system and the safety of its deployment. Consider Google Duplex \citep{Yaviv2018}, a TTS system for accomplishing real-world tasks over the phone. 
The \textit{inclusion of disfluencies} (such as fillers and repairs) led to highly natural sounding generated responses, showing how ubiquitous disfluencies are to everyday communication. However, these responses made human callers think that they were conversing with another human. This illusion of agency may have negative consequences when considering \textit{safety} in conversational AI. For example, attributing anthropomorphic traits to an agent in an \textsc{Impostor effect} \citep{safety,abercrombie2022risk} scenario, i.e. where a system provides inappropriate advice in safety critical situations (such as when a user seeks medical advice). Despite this, from  a generation perspective, fillers and disfluencies may still be desirable if used as a mitigation strategy, i.e. to exhibit uncertainty and doubt in safety critical situations (e.g. ``uh.. I'm not an expert here but...''). This is important, as \citet{mielke2022reducing} pointed out that neural dialogue agents are not linguistically calibrated -- i.e. despite the agent being \textit{factually inaccurate} it may still (inappropriately) verbalise expressions of confidence in responses.  Thus, \textit{the trade-off between naturalness and safety} merits consideration in future research, particularly when using disfluencies in generation tasks.

%% file: Conclusion.tex
\section{Conclusion}
\label{conclusion}

The preliminary aim of this article was to bring together several different perspectives on fillers that influence research in SLU. The article was motivated by the increasing popularity of voice assistant technologies, where often, the corpora used will invariably contain these spontaneous speech events. Yet, to the best of our knowledge, there was not an introductory source on these specific phenomena, that offered a \textit{breadth} of different perspectives and approaches -- particularly targeted towards the SLU and Conversational AI community. We were motivated to introduce these different perspectives, without exhaustively going into details that may overwhelm the reader. 

Our aim was to discuss these perspectives in a holistic way, that is considering them at various stages of the SLU pipeline; from underlying (pyscho)linguistic theory, to annotation, ASR and SLU perspectives, and finally, looking briefly at research on the generation of these phenomena. We have tried to the best of our ability to synthesise the main perspectives in an approachable way, and suggest further reading and other sources when appropriate. Additionally in each section we pinpoint (what we believe) are the main challenges and trends of each area. To do so, we identified two areas of research, i.e. psycholinguistic and computational perspectives. In the former, we focused on the research available on the production and comprehension of disfluencies, particularly fillers. In the computational approaches, we discussed the treatment of disfluencies by distinguishing between SLU as it is typically considered for SDS, and broader SLU tasks more concerned with social communication. We try to highlight the commonalities regarding both perspectives as well, i.e. discussing works in both areas that find disfluencies informative, consider disfluencies as noise, consider the listener's perception \dots . Looking forward, researchers might consider the balance between removing disfluencies to aid the robustness of a system versus including them as speech events that could offer social context, the trade-off between how natural sounding the system is versus how safe it may be in deployment, the integration of fine-grained context, and lastly, the taxonomy of such phenomena in order to make the research on them more accessible.